\pdfoutput=1

\documentclass[11pt]{article}

\usepackage[final]{acl}
\usepackage{times}
\usepackage{latexsym}

\usepackage[T1]{fontenc}

\usepackage[utf8]{inputenc}

\usepackage{microtype}

\usepackage{inconsolata}

\usepackage{graphicx}

\usepackage{microtype}
\usepackage{hyperref}
\usepackage{url}
\usepackage{booktabs}
\usepackage{graphicx}
\usepackage{multirow}
\usepackage{array}
\usepackage{subcaption}
\usepackage{longtable}
\usepackage{marvosym}
\usepackage{amsmath,amsfonts,bm}
\usepackage{subcaption}
\usepackage{stfloats}

\usepackage{tabularx}
\usepackage{caption}

\title{Post-hoc Study of Climate Microtargeting on Social Media Ads with LLMs: Thematic Insights and Fairness Evaluation}

\author{Tunazzina Islam \\
  Department of Computer Science \\
  Purdue University \\
  West Lafayette, IN 47907 \\
  \texttt{islam32@purdue.edu} \\\And
  Dan Goldwasser \\
  Department of Computer Science \\
  Purdue University \\
  West Lafayette, IN 47907 \\
  \texttt{dgoldwas@purdue.edu} \\}

\begin{document}
\maketitle
\begin{abstract}
Climate change communication on social media increasingly employs microtargeting strategies to effectively reach and influence specific demographic groups. This study presents a \textit{post-hoc} analysis of microtargeting practices within climate campaigns by leveraging large language models (LLMs) to examine Meta (previously known as Facebook) advertisements. Our analysis focuses on two key aspects: \textbf{demographic targeting} and \textbf{fairness}.
We evaluate the ability of LLMs to accurately predict the intended demographic targets, such as gender and age group.
Furthermore, we instruct the LLMs to generate explanations for their classifications, providing transparent reasoning behind each decision. These explanations reveal the specific thematic elements used to engage different demographic segments, highlighting distinct strategies tailored to various audiences. Our findings show that \textbf{\textit{young adults}} are primarily targeted through messages emphasizing \textit{activism and environmental consciousness}, while \textbf{\textit{women}} are engaged through themes related to \textit{caregiving roles and social advocacy}.
Additionally, we conduct a comprehensive fairness analysis to uncover biases in model predictions. We assess disparities in accuracy and error rates across demographic groups using established fairness metrics such as Demographic Parity, Equal Opportunity, and Predictive Equality. Our findings indicate that while LLMs perform well overall, certain biases exist, particularly in the classification of \textbf{\textit{male}} audiences. The analysis of thematic explanations uncovers recurring patterns in messaging strategies tailored to various demographic groups, while the fairness analysis underscores the need for more inclusive targeting methods. This study provides a valuable framework for future research aimed at enhancing transparency, accountability, and inclusivity in social media-driven climate campaigns. 

\end{abstract}

\section{Introduction}
Climate change represents one of the most pressing global challenges of the 21st century, necessitating widespread public awareness and engagement to drive meaningful environmental action \cite{moritz2013future,dessler1995science}. As traditional media channels evolve, social media has emerged as a pivotal platform for climate communication, enabling organizations, activists, and policymakers to disseminate information, mobilize support, and influence public discourse on environmental issues \cite{nosek2020fossil,hestres2017internet,adger2003adaptation}. The interactive and targeted nature of social media advertising allows for the customization of messages to resonate with specific audiences, thereby enhancing the effectiveness of communication strategies aimed at fostering climate awareness and behavioral change \cite{bloomfield2019circulation,walter2018echo,stoddart2016canadian}.

In recent years, the utilization of microtargeting strategies in social media campaigns has gained significant traction \cite{eldar2025political,islam2025aaai,nistor2024thinking}. Microtargeting involves the precise tailoring of messages to distinct demographic segments based on factors such as age, gender, location, and interests \cite{islam2025understanding,prummer2020micro,hersh2015,barbu2014advertising}. This approach leverages vast amounts of user data to craft personalized content that is more likely to engage and persuade individual users. In the context of climate communication, microtargeting can enhance the relevance and impact of messages, potentially leading to greater public engagement and support for environmental initiatives. However, the sophistication of these strategies also raises critical questions about the transparency, ethical implications, and overall effectiveness of targeted climate messaging \cite{islam2024uncoveringthm,islam2023analysis}.

\begin{figure}[h]
  \centering  
  \includegraphics[width= .5\textwidth]{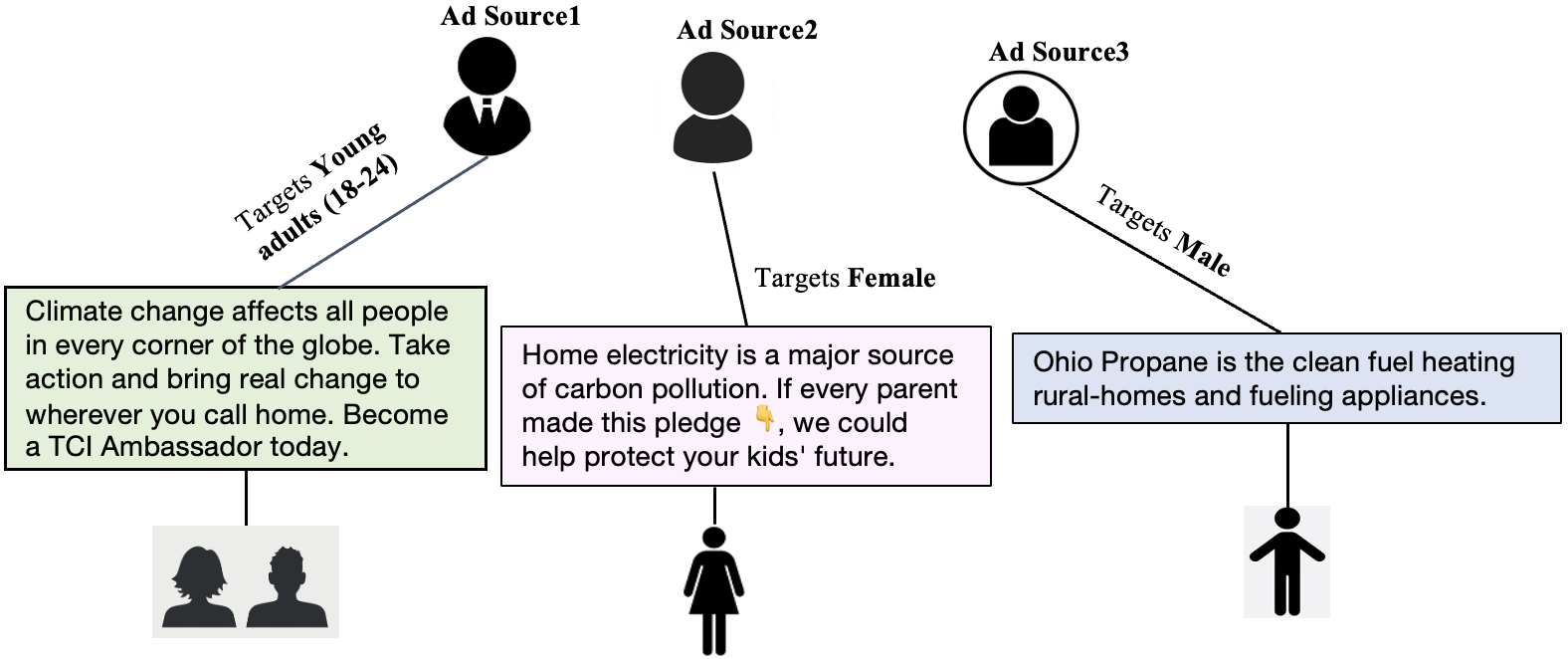}
  \caption{Example of climate microtargeting.}
  \vspace{-10 pt}
    \label{micro}
\end{figure}
Figure \ref{micro} illustrates the targeted climate advertisements on social media, with a specific focus on demographic targeting. The first ad source targets \textbf{young adults aged 18-24}, with a message \textit{encouraging action against climate change and inviting them to become The Climate Initiative (TCI)\footnote{https://theclimateinitiative.org/}) Ambassadors}. The second ad source is tailored to a \textbf{female} audience with a message emphasizing the \textit{importance of reducing carbon pollution from home electricity and making a pledge for their children's future}. The third ad source targets a \textbf{male} audience, focusing on the \textit{benefits of using clean fuel like Ohio Propane for heating rural homes and fueling appliances}. 


Despite the growing prevalence of microtargeting in climate campaigns, there remains a limited understanding of the specific techniques and linguistic patterns employed to engage different demographic groups. 
Traditional method \cite{roberts2019attempting,vaismoradi2016theme,vaismoradi2013content,braun2012thematic,braun2006using,tuckett2005applying} often falls short in capturing the nuanced and context-dependent nature of targeted communication. This gap highlights the need for advanced analytical tools that can dissect and interpret the complex language and strategies used in microtargeted climate advertisements. Large language models (LLMs) \cite{brown2020language}, with their robust natural language processing (NLP) capabilities, offer a promising solution to this challenge. 
In this paper, we investigate whether the newly emerging paradigm in NLP- zero-shot prompting of LLMs \cite{brown2020language} and the practice of providing explanations of answers are better equipped to address those challenges.
\begin{figure*}[h]
    \centering
	\includegraphics[width=\textwidth]{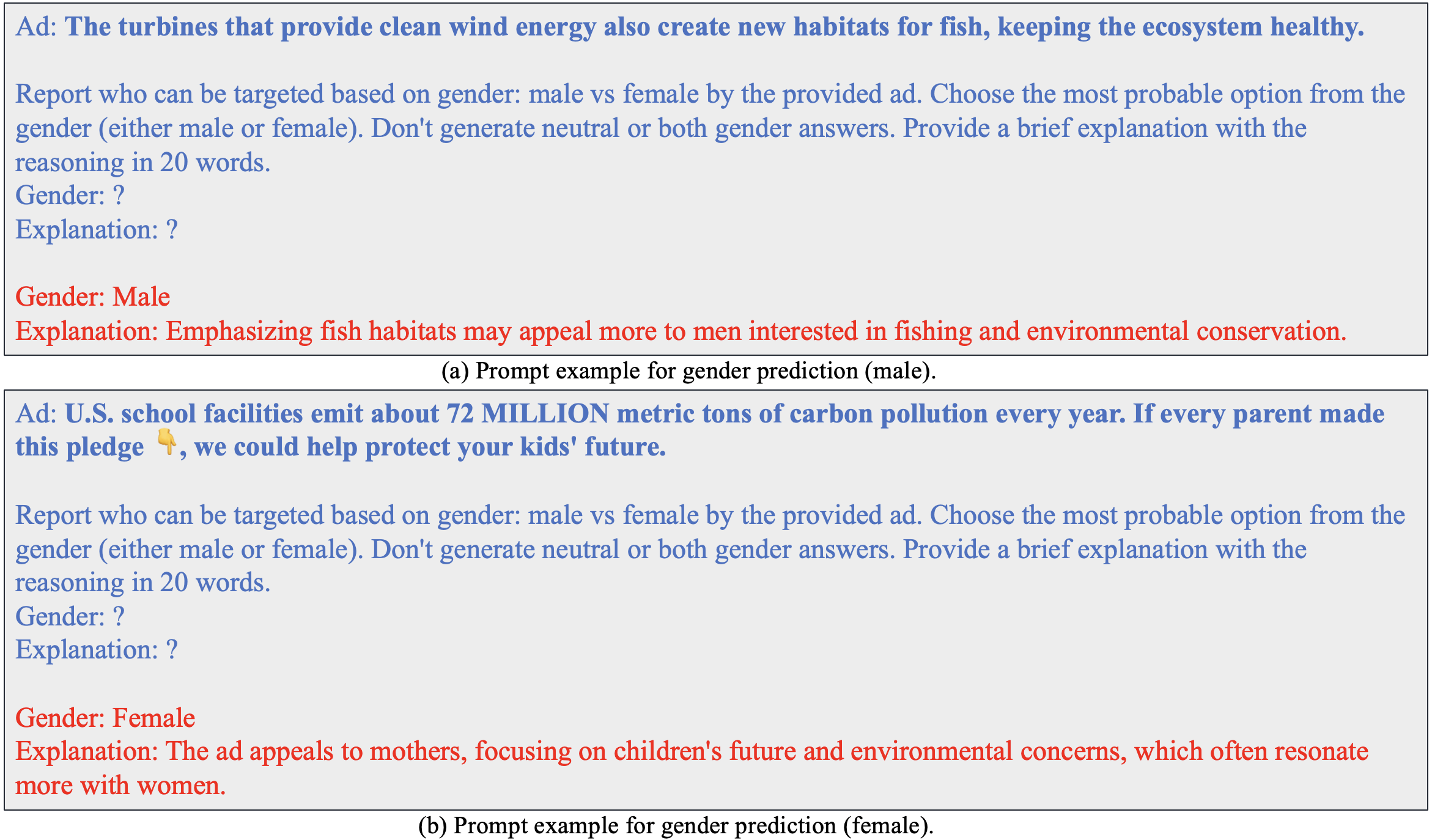}
	\caption{Prompt examples for gender prediction. (a) male, (b) female. Inputs are shown in blue, outputs in red.}
	\label{prompt_gen}
\end{figure*}

 Explanations are fundamental to human learning \cite{ahn1992schema}, as they underscore task principles that facilitate broad generalizations \cite{lombrozo2006functional,lombrozo2006structure}. Consider the example text ( {\small\textit{``The turbines that provide clean wind energy also create new habitats for fish, keeping the ecosystem healthy." }}) from Figure \ref{prompt_gen}(a) for the gender identification task. 
 An explanation can elaborate on a brief answer (e.g., {\small\texttt{\textbf{male}}}) by connecting it to the broader reasoning process necessary to solve the problem (e.g., “{\small\texttt{Emphasizing fish habitats may appeal more to men interested in fishing and environmental conservation.}}”).
 Thus, explanations enhance understanding by demonstrating how task principles connect questions to their answers.

While LLMs offer powerful capabilities for analyzing and generating text, their widespread use has also highlighted significant challenges related to fairness \cite{kumar2023language,li2023fairness,li2023survey,sag2023fairness} and bias \cite{lin2024investigating,kotek2023gender,fang2024bias,urman2023silence,esiobu2023robbie}. Research has shown that LLMs, like other machine learning (ML) models, can inherit and even amplify biases present in the data on which they are trained \cite{blodgett2020language,binns2018fairness}. These biases can manifest in various forms, such as differential accuracy across demographic groups \cite{dacon2021does,lu2020gender,hitti2019proposed,diaz2018addressing}, harmful stereotypes \cite{bianchi2023easily,matthews2022gender,alemany2022methodology}, and discriminatory language patterns \cite{salguero2018flexible}. The implications are particularly concerning in high-stakes applications such as healthcare, finance, and social networks, where biased outcomes can perpetuate inequities and undermine trust in artificial intelligence (AI) systems \cite{mehrabi2021survey,buolamwini2018gender}. Addressing these issues requires fairness evaluations using metrics like Demographic Parity, Equal Opportunity, and Predictive Equality to ensure that models perform equitably across all user groups \cite{vzliobaite2017measuring,hardt2016equality}. In the context of this study, examining the fairness of LLMs' predictions in demographic targeting is crucial for understanding the microtargeted climate messaging.

In this study, we conduct a \textit{post-hoc} analysis of climate microtargeting practices on social media by leveraging the power of LLMs (\textbf{OpenAI's o1-preview model}\footnote{https://openai.com/index/learning-to-reason-with-llms/}). Post-hoc analysis, typically performed after the main experiment or event, allows us to retrospectively evaluate how effective these campaigns are in targeting specific demographics. Building upon data from previous research by \citet{islam2023analysis,islam2024uncoveringthm}, we investigate the ability of LLMs to accurately detect targeted messaging based on specific demographic variables, including gender and age group. Additionally, LLMs provide explanations for their classification decisions, offering insights into the thematic and linguistic elements used to engage different audiences. Furthermore, we conduct a comprehensive fairness analysis to identify potential biases in model predictions. 
Our exploration leads to the following research questions (RQ), which are crucial for assessing the potential of LLMs to understand microtargeting patterns and provide deeper and more nuanced insights:
\begin{itemize}
    \item \textbf{RQ1:} Given a text, can LLMs identify the targeted demographic of the corresponding text accurately and provide an explanation for the reasoning behind the prediction?
    \item \textbf{RQ2:} What are the recurring themes and aspects of \textbf{explanations} provided by LLMs?
    \item \textbf{RQ3:} How fair are the LLMs' predictions in terms of demographic targeting, and what are the disparities in prediction accuracy and error rates across different demographic groups?
\end{itemize}

The main idea of this paper is to utilize LLMs as a tool to analyze real-world targeting practices, measuring biases and fairness in actual campaigns. We explore how LLMs can be used post-hoc as third-party tools to analyze patterns in targeted communication, especially when \textbf{internal platform logic is not transparent}. While platforms have white-box access, external stakeholders (researchers, auditors, policymakers) \textbf{do not}. Our method offers an \textit{explainable} approach to \textit{reverse-engineer} targeting practices and uncover potential bias or messaging disparities.

The implications of this research are multifaceted. From an academic perspective, it contributes to the burgeoning field of computational social science (CSS) by showcasing the application of advanced language models in dissecting complex communication strategies. Practically, the findings offer valuable insights for policymakers, environmental organizations, and social media platforms seeking to enhance the transparency, accountability, and inclusivity of their climate communication efforts. By illuminating the specific methods used to tailor messages to different demographics and by highlighting the need for fairer and more inclusive targeting methods, this study lays the groundwork for future investigations into the role of AI in enhancing the efficacy and ethical standards of digital climate advocacy.
\section{Related Work}
The intersection of microtargeting, social media, and climate communication has drawn considerable interest in recent years, with studies examining how digital platforms influence public opinion and engagement \cite{islam2024uncovering,islam2024uncoveringthm,holder2023climate,islam2023analysis}. The growing capabilities of social media platforms to deliver personalized messages based on user data have sparked a significant body of research into computational advertising \cite{abrams2022targeting,huh2020advancing,yang2017computational,zhang2017targeted}, demographic targeting \cite{allen2022targeted,rummo2020examining,ribeiro2019inference,kaspar2019personally,jansen2013evaluating,jansen2010gender}, and the broader implications of these practices for societal discourse.

Microtargeting on social media platforms such as Meta has been widely studied in the context of political campaigns \cite{islam2023weakly,capozzi2021clandestino,serrano2020political,silva2020facebook,capozzi2020facebook} and public health messaging \cite{islam2022understanding,silva2021covid,mejova2020covid}, revealing both the potential benefits and ethical concerns associated with this practice. In the realm of climate communication, microtargeting can be a powerful tool for enhancing message relevance and impact by tailoring content to the specific values and interests of diverse demographic groups \cite{islam2023analysis}. However, the effectiveness and ethical implications of such targeted messaging remain underexplored, particularly in terms of how different demographic groups engage and whether biases are present in targeting strategies.


Recent advances in LLMs have demonstrated their capability for in-context learning (ICL), significantly enhancing their ability to perform tasks traditionally handled by humans \cite{chowdhery2023palm,kojima2022large,le2022bloom,brown2020language,gilardi2023chatgpt,de2023can,dai2023llm,chiang2023can,ziems2024can}. This progress suggests a strong potential for effectively applying LLMs to our specific task. In the realms of qualitative research (QR) and NLP, innovative methods are being explored to integrate LLMs into Thematic Analysis (TA) \cite{braun2012thematic}. Researchers have proposed various frameworks, including an LLMs-in-the-loop approach \cite{islam2024uncovering,islam2024uncoveringthm,dai2023llm}, integrating GPT-3 with expert-designed codebooks \cite{xiao2023supporting}, and developing collaborative interfaces that utilize LLMs for code generation and support in decision-making processes \cite{gao2023collabcoder}. Other recent work has shown that LLMs can benefit from examples that decompose the reasoning process (can be seen as an explanation), leading to an answer \cite{wei2022chain}. 
Despite the impressive capabilities of LLMs, there are concerns about fairness, accountability, and transparency in LLMs' predictions, which have been highlighted in recent literature \cite{anthis2024impossibility,dai2024bias,kumar2023language,li2023fairness,wu2023brief,bender2021dangers}, emphasizing the need for a rigorous evaluation of biases and disparities in model performance across different groups. 
In this paper, we leverage LLMs to identify targeted demographics and provide explanations of demographic targeting regarding climate-related advertisements on Meta.  Besides, we develop a new set of themes and aspects based on those explanations, specifically tailored for analyzing messaging. Additionally, we extend previous research by not only focusing on the accuracy of these predictions but also conducting a comprehensive fairness analysis to identify and address potential biases in the model's performance.
\section{Dataset}
We investigate the climate campaigns case study for this work. We work on the corpus of $21372$ ads \cite{islam2024uncoveringthm} originally released by \citet{islam2023analysis}. This dataset includes climate-related English ads on Meta from the US, spanning from January $2021$ to January $2022$.
Each ad includes the following attributes: ad ID, title, ad description, ad body, funding entity, spend, impressions, and distribution of impressions broken down by gender (male, female, unknown), age (seven groups), and location down to the state level in the USA. Additional details about the dataset can be found in the original publications. 

For this work, we consider two demographic indicators, i.e., \textbf{gender} and \textbf{age group}. We consider two gender categories, i.e., \textit{male} vs \textit{female}. Regarding age group, we consider four age group categories, i.e., \textit{young adults} whose age range is 18-24, \textit{early working} age group (25-44), \textit{late working} age group (45-64), and \textit{senior citizens} (65+). However, the Meta Ad Library API doesn't provide explicit targeting information beyond ad impressions (views). Note that this paper \textbf{infers targeting from exclusive impression distributions, which may differ from advertiser intent}.
In this work, we focus only on ads viewed exclusively by either males or females; ads viewed by both genders were excluded. Therefore, if an ad is viewed solely by males or solely by females, we consider it a targeted ad. The same logic is applied to age groups. We do not include any ads that overlap across age categories, such as young adults, early working-age individuals, late working-age individuals, and senior citizens. By following this methodology, we narrow down the dataset to $227$ ads. Among them, $59$ ads target only females and $47$ ads target only males. However, we find $25$ ads target only young adults, $82$ ads target only the early working age group, $8$ ads target only the late working age group, and $6$ ads target only senior citizens. 

\section{Experimental Setup}
The identification of the targeted demographic (with explanation) in a text
involves the following steps:
\newline
\textbf{Gender prediction with Explanation:} For a given text $t$, the task involves identifying the targeted gender and explaining the rationale behind its selection.
\newline
\textbf{Age group prediction with Explanation:} Subsequently, the task requires predicting the targeted age group and providing an explanation for the specific choice.

To identify the targeted demographics, we use the most recently\footnote{September 12, 2024} released by OpenAI, \textbf{o1-preview model}\footnote{https://openai.com/index/introducing-openai-o1-preview/}. This is a new, large language model that uses reinforcement learning (RL) and chain of thought (COT) techniques for complex reasoning, allowing it to think through a detailed internal process before responding to users. We use the OpenAI Playground API to run \textbf{o1-preview} by keeping the default parameters. The prompt template for the demographic prediction task with an explanation using LLMs can be found in Fig. \ref{prompt_tem} in App. \ref{app:prmpt}. Fig. \ref{prompt_gen} shows the example prompts for gender prediction. Fig. \ref{prompt_age} in App. \ref{app:prmpt} shows the example prompts for age group prediction from the climate campaign dataset.
\begin{table*}
\centering
\begin{tabular}{l|c|c|c|c}
\hline
\textbf{Category} & \textbf{Total Ads} & \textbf{Correct Pred.} & \textbf{Acc. (\%)} & \textbf{Misclass.} \\ \hline
All & 227 & 201 & 88.55 & - \\ \hline
Female & 59 & 56 & 94.92 & 3 (Male) \\ \hline
Male & 47 & 40 & 85.10 & 7 (Female) \\ \hline
Young adults & 25 & 22 & 88.00 & 2 (Early Working), 1 (Late Working) \\ \hline
Early Working & 82 & 75 & 91.46 & 4 (Young), 4 (Late Working) \\ \hline
Late Working & 8 & 6 & 75.00 & 2 (Early Working) \\ \hline
Senior & 6 & 2 & 33.33 & 3 (Young), 1 (Late Working) \\ \hline
\end{tabular}
\caption{Accuracy and Misclassifications for Demographics.}
\label{result}
\end{table*}
%
\subsection{Results}
Table \ref{result} provides the overall accuracy of the targeted demographic prediction task by LLMs as well as a detailed breakdown of correct and incorrect predictions across each demographic category. LLMs can predict the targeted demographics with an accuracy of $88.55\%$ answering \textbf{RQ1}.  
LLMs achieve high accuracy in predicting both females ($94.92\%$) and males ($85.10\%$). A small number of females are misclassified as males, and a few males are misclassified as females. Table \ref{result} shows high accuracy for young ($88.00\%$) and early working ($91.46\%$) categories. Performance drops for the late working ($75\%$) age group and significantly for the senior ($33.33\%$) categories. 

For baseline comparison, we use open-sourced LLMs Llama 3 (llama3-70b-8192\footnote{\url{https://github.com/meta-llama/llama3}}) \cite{touvron2023llama} and Mistral Large 2 (mistral-large-2407\footnote{\url{https://mistral.ai/news/mistral-large-2407/}}) \cite{jiang2023mistral}. For llama3-70b-8192 and mistral-large-2407, we keep their default parameters. To run llama3-70b-8192, we use Groq API\footnote{\url{https://wow.groq.com/}}. For the tokenizer, we use Meta-Llama-3-70B-Instruct from Hugging Face. For running Mistral Large 2, we use the Mistral AI API.  Besides, we include a smaller pre-trained language model BERT \cite{devlin-etal-2019-bert} (bert-base-uncased) for comparison.
Furthermore, we compare with simple Logistic Regression (LR) \cite{cox1958regression} trained on term frequency–inverse document frequency (tf-idf) features baseline. However, OpenAI's o1-preview model outperforms the baselines both in gender (macro average F1 score $90.35\%$) and age group (macro average F1 score $71\%$) predictions (Table \ref{baseline}).
\begin{table}[h]
\centering
\begin{tabular}{l|l|c|c}
\toprule
\textbf{Model} & \textbf{Demo.} & \textbf{Acc. (\%)} & \textbf{F1 (\%)} \\
\midrule
\multirow{2}{*}{LR\textsubscript{tf-idf}} & gender & 69.00 & 65.00 \\
& age    & 73.00 & 31.00 \\
\hline
\multirow{2}{*}{BERT} & gender & 72.00 & 70.00 \\
& age    & 70.00 & 26.00 \\
\hline
\multirow{2}{*}{Llama 3} & gender & 80.19 & 79.67 \\
& age    & 58.68 & 36.84 \\
\hline
\multirow{2}{*}{Mistral Large 2} & gender & 82.08 & 82.07 \\
& age    & 74.38 & 48.68 \\
\hline
\multirow{2}{*}{\textbf{o1-preview}} & \textbf{gender} & \textbf{90.57} & \textbf{90.35} \\
& \textbf{age}    & \textbf{85.95} & \textbf{71.00} \\
\bottomrule
\end{tabular}
\caption{Baseline comparisons. Demo.: Demographics, Acc.: Accuracy.}
\vspace{-10 pt}
\label{baseline}
\end{table}

\begin{table*}
\centering
\begin{tabular}{>{\raggedright\arraybackslash}p{.8 in}|>{\raggedright\arraybackslash}p{1.8 in}|>{\raggedright\arraybackslash}p{2.8 in}}
\hline
\textbf{Gender} & \textbf{Theme of Explanation} & \textbf{Aspects of Explanation} \\ \hline
Male & Perceived Interests and Roles & 
     Interest in Technology and Innovation, Focus on Economic and Financial Issues, Property and Land Management, Traditional Male Activities, Engagement in Political and Infrastructure Topics, Conservative Views and Skepticism\\ 
\hline
Female & Roles as Caregivers, Environmental Advocates, and Socially Conscious Individuals & 
    Parental and Caregiving Roles, Environmental Consciousness, Social Welfare and Community Involvement, Empathy and Emotional Appeal, Female Empowerment and Leadership, Health and Safety Concerns \\
\hline
\end{tabular}
\caption{Gender based Themes and Aspects of Explanations.}
\label{gender_thm}
\end{table*}
\subsection{Statistical Tests}
We perform a binomial test assessing whether the model's gender predictions are significantly better than random guessing ($50\%$). 
\newline
Null hypothesis, $H_o$ : probability of correct prediction is $0.5$ (random guessing), p-value: $\sim 4.37 \times 10^{-19}$, Conclusion: Reject $H_o$. 
This means the model's accuracy is significantly better than chance, with a very strong statistical significance. Table \ref{subgen_stat} shows the statistical test results from the gender subgroup.
\begin{table}[h!]
\centering
\begin{tabular}{l|c|c}
\toprule
\textbf{Gender} & \textbf{p-value} & \textbf{Conclusion} \\
\midrule
Male   & $5.35 \times 10^{-7}$  & Reject H$_0$ \\
Female & $5.95 \times 10^{-14}$ & Reject H$_0$ \\
\bottomrule
\end{tabular}
\caption{Significance tests by subgroup (gender).}
\label{subgen_stat}
\end{table}

Here are the results of the binomial test for age group predictions, assuming random guessing across $4$ categories ($p = 0.25$): 
\newline
Null hypothesis, $H_o $: probability of correct prediction is $0.25$ (random guessing), p-value: $\sim 4.27 \times 10^{-45}$, Conclusion: Reject $H_o$. The model’s performance is significantly better than random chance. This confirms that the predictions are statistically meaningful. Table \ref{subage_stat} shows the statistical test results from the age subgroup.
\begin{table}[h!]
\centering
\begin{tabularx}{\columnwidth}{X|X|X}
\toprule
\textbf{Age Group} & \textbf{p-value} & \textbf{Conclusion} \\
\midrule
Senior        & $4.66 \times 10^{-1}$  & Fail Reject H$_0$ \\
EarlyWorking & $1.04 \times 10^{-35}$ & Reject H$_0$ \\
Young         & $5.76 \times 10^{-11}$ & Reject H$_0$ \\
Late Working  & $4.23 \times 10^{-3}$  & Reject H$_0$ \\
\bottomrule
\end{tabularx}
\caption{Significance tests by subgroup (age).}
\label{subage_stat}
\end{table}

\subsection{Error Analysis}
Table \ref{error} in App. \ref{app:error} presents an analysis of ad misclassifications based on gender and age group predictions. Each entry includes the actual demographic, the predicted demographic, and a brief explanation generated by LLMs. Explanations highlight how specific patterns and themes within an ad can lead to demographic misclassifications. In some cases, \textbf{traditional gender roles} and \textbf{age-related interests }played a significant role in these misclassifications. Understanding these nuances can help in refining predictive models and improving the accuracy of demographic targeting in future ad campaigns.
\begin{table*}
\centering
\begin{tabular}{>{\raggedright\arraybackslash}p{0.8in}|>{\raggedright\arraybackslash}p{1.8in}|>{\raggedright\arraybackslash}p{2.8in}}
\hline
\textbf{Age group} & \textbf{Theme of Explanation} & \textbf{Aspects of Explanation} \\ \hline
Young adults (18-24) & Activism and Environmental Consciousness & 
     Passion for Climate Action, Support for Bold Environmental Leadership,   Engagement with Activism, Desire for Immediate Change, Participation in Training and Advocacy \\
\hline
Early working (25-44) & Proactive and Responsible Mindset & 
    Environmental Consciousness,
     Financial Stability and Disposable Income,
     Parental and Future Concerns,
    Career Engagement and Professional Roles,
    Interest in Innovation and Technology,
    Social and Political Engagement \\
\hline
Late working (45-64) & Responsibilities and Concerns & 
    Economic and Environmental Responsibility,
     Homeownership and Financial Stability,
     Voter and Policy Engagement,
    Economic Concerns \\ 
\hline
Senior (65+) & Health and Safety Concerns & 
   Health and Wellness Programs,
   Vulnerability and Safety \\ 
\hline
\end{tabular}
\caption{Age group based Themes and Aspects of Explanations.}
\label{age_thm}
\end{table*}
\section{Thematic Insights of Explanations}
As LLMs provide explanations to provide the reasoning behind their prediction, we use those explanations to understand thematic insights to answer \textbf{RQ2}. In this analysis, we \textbf{only include the correct predictions and their explanations}. Using only correct predictions is \textbf{not arbitrary}—it is supported by \textbf{methodological norms} \citep{vaismoradi2016theme,vaismoradi2013content,braun2012thematic,braun2006using} to ensure pattern coherence. 
\subsection{Themes and Aspects of Gender Explanations}
We prompt LLMs to provide the common theme and aspects under the specific theme of the explanations from $40$ correct \textbf{male} predictions and $56$ correct \textbf{female} predictions. The prompt template is shown in Figure \ref{thm_gen} App. \ref{app:prmpt}. We detail the theme of gender explanation and the aspects of the explanation in Table \ref{gender_thm}. Two experts in NLP and CSS meticulously evaluate the generated themes and aspects derived from the explanations, ensuring their accuracy and alignment with the intended context. The inter-annotator agreement is $0.94$ (almost perfect agreement) using Cohen’s Kappa coefficient \cite{cohen1960coefficient}. We detail the theme of the gender explanations and aspects of that explanation in App. \ref{app:thm_gen}.
\subsection{Themes and Aspects of Age Explanations}
We prompt LLMs to provide the common theme and aspects under the specific theme of the explanations from $22$ correct \textbf{young adult} predictions, $75$ correct \textbf{early working age group} predictions, $6$ correct \textbf{late working age group} predictions, and $2$ correct \textbf{senior citizen} predictions. We detail the theme of the age group explanation and aspects in Table \ref{age_thm}. Similarly to gender, the same two NLP and CSS experts review the generated themes and aspects from the explanations to ensure accuracy and contextual alignment. For the age group, the inter-annotator agreement is $0.91$ (almost perfect agreement) using Cohen’s Kappa coefficient.
We detail the theme of the age group explanations and aspects of that explanation in App. \ref{app:thm_age}.
\subsection{Baselines}
We add Latent Dirichlet Allocation (LDA) \cite{blei2003latent} and BERTopic \cite{grootendorst2022bertopic} as unsupervised techniques for generating topics from LLMs-generated explanations. BERTopic\footnote{\url{https://maartengr.github.io/BERTopic/\#common}} is a hierarchical topic modeling and an off-the-shelf neural baseline for clustering texts.
BERTopic generates two topics from the prediction explanation of gender and age (Table \ref{tab:topic_modeling_results} in App. \ref{unsup_base}). We run LDA with four topics and show the top $10$ words in App. \ref{unsup_base} in Table \ref{tab:topic_modeling_results}. We perform a human evaluation by asking annotators to assess whether the generated themes and topics match the ad's explanation. Table \ref{tab:human} in App. \ref{unsup_base} shows the human evaluation results on LLM (GPT-4o \cite{openai2024gpt4o}) assignment, LDA, and BERTopic.
\section{Fairness and Bias Analysis}
In this section, we present a comprehensive fairness analysis of the model for gender and age group classifications to answer \textbf{RQ3}. We evaluate the models using established fairness metrics such as Demographic Parity, Equal Opportunity, and Predictive Equality to assess their performance across different groups. By analyzing confusion matrices and classification reports, we identify any disparities in prediction accuracy and error rates between groups. Our analysis aims to identify biases and investigate the underlying reasons for any observed biases. The insights gained from this analysis are critical for guiding future research in developing fair and inclusive algorithms.
%
\subsection{Fairness Analysis on Gender Prediction}
To assess the fairness of the model on gender prediction, we compute several fairness metrics, including Demographic Parity, Equal Opportunity, and Predictive Equality.
\newline
\textbf{Demographic parity} examines whether each gender group receives positive predictions at equal rates. Table \ref{demo_parity_gen} presents the Demographic Parity ratios for each gender. A ratio of $1$ indicates perfect parity. The results show that \textbf{females have a slightly higher likelihood of receiving positive predictions compared to males}, suggesting a minor imbalance favoring the female class.
\newline
\textbf{Equal Opportunity (True Positive Rate)} focuses on the True Positive Rates (TPR) across gender groups, measuring the model's ability to correctly identify positive instances within each group. We have \textbf{Female TPR:} $\mathbf{0.95}$ and \textbf{Male TPR:} $\mathbf{0.85}$. The TPR for females is higher by $0.10$, indicating that the model may be more effective at correctly identifying females than males.
\newline
\textbf{Predictive Equality (False Positive Rate)} assesses the False Positive Rates (FPR) across gender groups, reflecting the rate at which negative instances are incorrectly labeled as positive. We achieve \textbf{Female FPR:} $\mathbf{0.07}$ and \textbf{Male FPR:} $\mathbf{0.05}$. The slightly higher FPR for females suggests that \textbf{females are more likely to be incorrectly predicted as positive compared to males}.
\begin{table}[h!]
    \centering
    \begin{tabular}{l|c}
        \toprule
        \textbf{Gender} & \textbf{Demographic Parity Ratio} \\
        \midrule
        Female & 1.0678 \\
        Male & 0.9149 \\
        \bottomrule
    \end{tabular}
    \caption{Demographic Parity for Predicted Gender.}
    \vspace{-5 pt}
    \label{demo_parity_gen}
\end{table}
\begin{table}[h!]
    \centering
    \begin{tabular}{l|c}
        \toprule
        \textbf{Age Group} & \textbf{Demographic Parity Ratio} \\
        \midrule
        Early Working & 0.95\\
        Late Working  & 1.50 \\
        Senior        & 0.33 \\
        Young         & 1.16 \\
        \bottomrule
    \end{tabular}
    \caption{Demographic Parity for Predicted Age Group.}
    \vspace{-5 pt}
    \label{demo_parity_age}
\end{table}
\begin{table}[h!]
    \centering
    \begin{tabularx}{\columnwidth}{X|X|X}
        \hline
        \textbf{Age Group} & \textbf{Equal Opportunity (TPR)} & \textbf{Predictive Equality (FPR)} \\
        \hline
        Early Working & 0.90 & 0.10 \\
        Late Working & 0.75 & 0.05 \\
        Senior & 0.33 & 0.00 \\
        Young & 0.88 & 0.07 \\
        \hline
    \end{tabularx}
    \caption{TPR and FPR for Age Groups.}
    \vspace{-10 pt}
    \label{equal_opp_pred_equal}
\end{table}
\subsection{Fairness Analysis on Age Prediction}
Results in Table \ref{demo_parity_age} show that the \textbf{late working group has a higher likelihood of receiving positive predictions}, while the \textbf{senior group has a significantly lower rate,} indicating \textbf{potential bias} against \textbf{seniors}. The senior group has a Demographic Parity ratio of $0.33$, significantly lower than the ideal value of $1$, suggesting \textbf{under representation in positive predictions}.

Table \ref{equal_opp_pred_equal} shows that the TPR for the senior group is notably lower, suggesting the model is less effective at correctly identifying individuals in this age group. The senior group has an FPR of $0$, indicating no false positives, while the early working group has the highest FPR among the groups. 

To understand the observed bias in the misclassified senior age group, we conduct an analysis of the misclassified instances (Table \ref{mis_class_senior} in App. \ref{app:senior}). The goal is to identify patterns and underlying reasons why the model predicts seniors as belonging to the young and late working age group. The misclassification of the Senior age group as Young or Late Working can be attributed to several factors, such as: lack of age-specific cues, reliance on stereotypical associations, feature representation limitations etc. We detail in App. \ref{app:senior}.
\subsection{Statistical Significance Tests}
While $0.10$ difference in TPR may seem modest, prior fairness literature \cite{hardt2016equality} recognizes it as meaningful in sensitive domains. We conduct two-proportion Z-tests to assess whether the observed differences in TPRs across demographic groups are statistically significant.
For gender prediction, the difference in TPR between females ($0.95$) and males ($0.85$) resulted in a p-value of $0.086$ with a $95\% $ confidence interval of $[-0.018, 0.230]$. This result indicates that the gap is not statistically significant at the conventional $0.05$ level. However, in the age group analysis, the TPR difference between young adults ($0.88$) and seniors ($0.33$) yields a p-value of $0.004$ with a $95\%$ confidence interval of $[0.138, 0.796]$, indicating a statistically significant disparity. This validates our concern about fairness in age-related predictions, particularly for underrepresented senior groups. 
\subsection{Sensitivity Analysis}
Our study is descriptive and \textbf{does not claim causal effects}. We have added sensitivity analysis (Fig. \ref{fig:sense} in App. \ref{senstv_ana}) by rerunning the fairness test after stratifying by \textit{ad-spend} and by \textit{topic cluster} to confirm the reported biases persist under two plausible confounders. 

Fig \ref{fig:gps} in App. \ref{senstv_ana} shows that bias persists after conditioning on ad spend. Low-budget ads: $\Delta \approx 0.82 $ ($95\%$ CI $0.68$ – $0.96$), High-budget ads:  $\Delta \approx 0.76 $ ($95\%$ CI $0.55$ – $0.93$). Overlapping CIs indicate similar magnitude, and \textbf{ad-budget is not driving the disparity}. Topic clusters tell the same story (Fig. \ref{fig:gpt} in App. \ref{senstv_ana}). Four of five clusters show sizable positive gaps ($0.30$ – $1.00$). The one cluster with only female-authored ads (topic 2) cannot yield a gap, but its omission does not overturn the pattern. Accuracy gaps are modest and hover around zero. Where CIs cross zero, classification accuracy is comparable between genders; the \textbf{selection-rate bias remains the dominant concern}.

Fig \ref{fig:aps} in App. \ref{senstv_ana} illustrates that the gap is not a low-budget artifact: whether an ad costs $\$5$ or $\$500$, the model is still far more likely to assign it to some age groups than others. Four of five topics show large positive parity gaps—even where sample sizes differ. In Fig. \ref{fig:apt} in App. \ref{senstv_ana}, Topic 1 does not contradict the bias claim as it simply contains ads that all target (and therefore are predicted for) one single age group. Stratified sensitivity checks confirm that age-group disparities persist across both advertising-budget tiers ($\Delta \approx 0.47 $ – $1.0$) and four of five latent topic clusters, indicating that \textbf{neither spend nor topic accounts for the bias}.

\subsection{Robustness Analysis}
To provide a more in-depth evaluation of the quality and robustness of the data, we plot a learning curve containing macro-F1 and demographic-parity against sample size ($20$–$227$ ads). Learning-curve plot in Fig. \ref{fig:lcr} in App. \ref{lc_robust} shows that macro-F1 and fairness gaps plateau well before 227 examples—evidence that the \textbf{analysis is not variance-limited}. 

Both curves (Fig. \ref{fig:lc_f1_gen} and Fig. \ref{fig:lc_fair_gen} App. \ref{lc_robust}) plateau by $\approx 60$ examples, confirming that our results are robust to subsampling and that additional data in the current distribution yield diminishing returns. Both performance and bias saturate after $\sim 1/4 $ of the data, indicating the analysis is not variance-limited.

Figures \ref{fig:lc_f1_age} and \ref{fig:lc_fair_age} in App. \ref{lc_robust} show the macro-F1 and age-parity gap at 20-ad increments. Macro-F1 rises to $0.70$ by $n \approx 80$ and then plateaus; the $95\%$ confidence band narrows to $±0.8$ pp. The parity gap likewise converges to $\approx 0.57$ with error $< 0.02$. Thus, our age-bias findings are not driven by sample variance—adding more data in the current distribution neither boosts accuracy nor mitigates disparity.

\section{Conclusion}
This study shows that LLMs can function as practical, third-party auditors of microtargeted climate ads—identifying \textit{whom messages are meant to reach}, \textit{explaining why}, and surfacing \textit{equity concerns} that are otherwise opaque. 
On our climate-ad corpus, the LLMs accurately infer demographic targets overall ($88.55\%$ accuracy), with especially strong performance for gender ($94.92\%$ for female-targeted ads; $85.10\%$ for male-targeted ads) and more variable results across age groups.
Beyond prediction, model-generated explanations enable a compact, reusable taxonomy of themes used to engage different audiences, providing interpretable signals that researchers and practitioners can inspect and act on. 
Besides, the fairness analysis conducted in our study underscores the importance of evaluating and addressing the biases. Disparities in prediction accuracy and error rates highlight the need for more inclusive and equitable targeting methods. 
Although we show our analysis on the climate campaigns dataset, our approach can easily be adapted to any dataset. It is designed to be scalable without any modifications. Code and dataset are available\footnote{\url{https://github.com/tunazislam/llms-posthoc-climate}}.
\section{Limitations}
Though we show the climate microtargeting case study, it can be applicable to political or health-related campaigns. However, the main idea of this work is to utilize LLMs as a tool to analyze real-world targeting practices, measuring biases and fairness in actual campaigns.

Our analysis relies on the OpenAI o1-preview model. We chose o1-preview instead of the open-source counterparts due to computational resource constraints. We only used pre-trained LLMs and did not consider fine-tuning due to the resource constraints. 

The original dataset does not specify intersectional targeting, such as young females only. Our paper acknowledges that our approach infers targeting from exclusive impression data (view), since the Meta Ad Library API does not disclose actual targeting parameters. Meta's own transparency studies treat ads whose impressions are $\geq 95\%$ in one group as ``effectively exclusive". Such ads deliver the same real-world exposure as explicit targeting. 

We have a small sample size due to restrictive filtering criteria, which might result in underrepresented groups, particularly late working and senior demographics. The rationale for using this specific dataset is that it is publicly available and, to the best of our knowledge, it is the only publicly available dataset focused on targeted climate campaigns.

%

\section{Ethics Statement}
To the best of our knowledge, we did not violate any ethical code while conducting the research work described in this paper. We report the technical details for the reproducibility of the results.
%
In this paper, we did not introduce any new dataset; instead, we experimented using an existing dataset that is adequately cited. The data do not contain personally identifiable information and report engagement patterns at an aggregate level.
The author's personal views are not represented in any qualitative results we report, as it is solely an outcome derived from machine learning or AI models.

\section{Acknowledgments} 
We are grateful to the anonymous reviewers for their insightful comments. 
This work was partially funded by the Purdue Graduate School Summer Research Grant (to TI) and NSF CAREER award IIS-2048001.


\bibliography{custom}

\appendix

\section{Appendix}
\label{sec:appendix}
\subsection{Prompting}
\label{app:prmpt}
\begin{figure*}
    \centering
	\includegraphics[width=\textwidth]{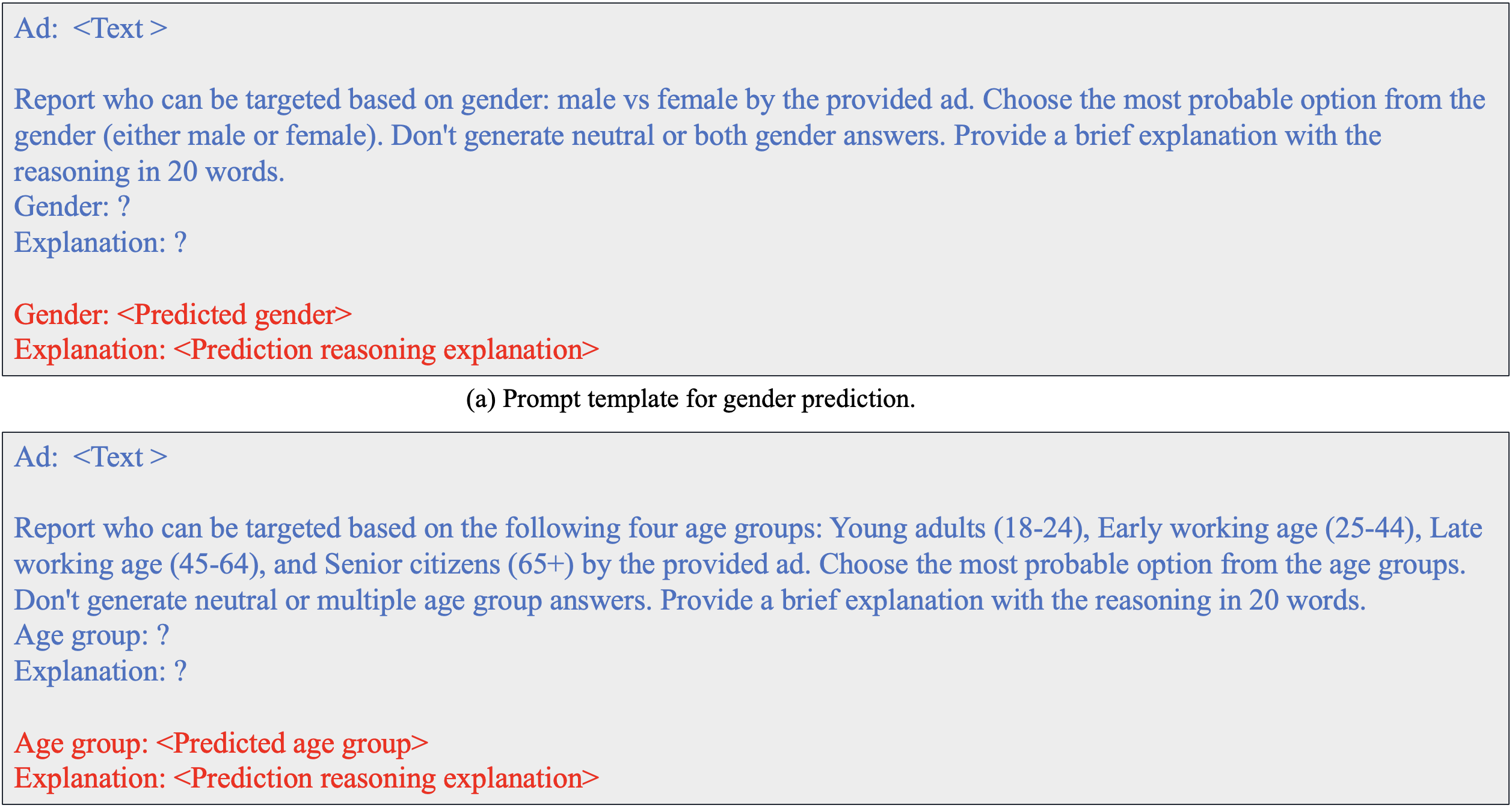}
	\caption{Prompt template for targeted demographic prediction. (a) gender, (b) age group. Inputs are shown in blue, and outputs are shown in red.}
	\label{prompt_tem}
\end{figure*}

\begin{figure*}
    \centering
	\includegraphics[width=\textwidth]{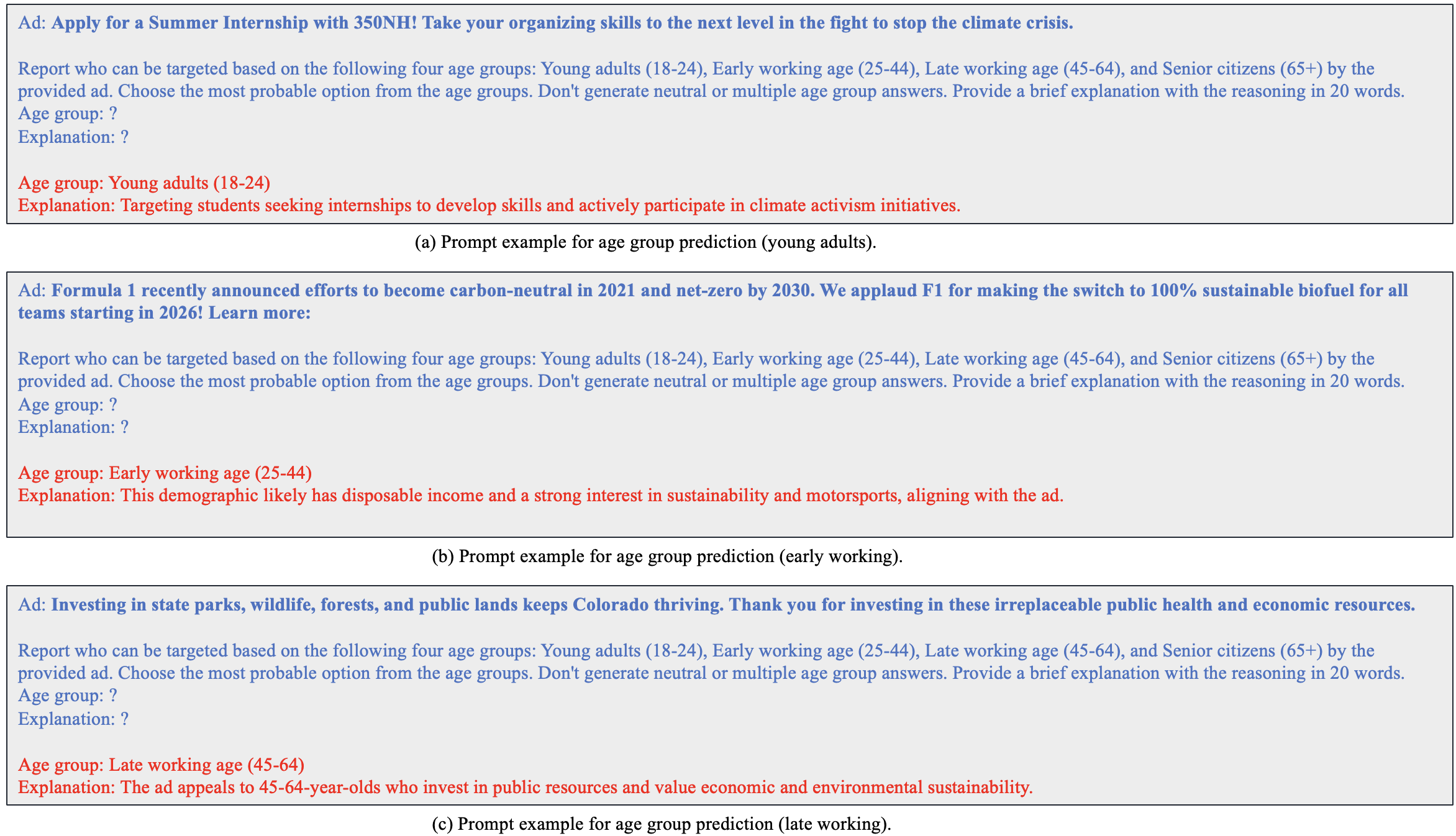}
	\caption{Prompt examples for age group prediction. (a) young adults, (b) early working age, (c) late working age. Inputs are shown in blue, and outputs are shown in red.}
	\label{prompt_age}
\end{figure*}

\begin{figure*}
    \centering
	\includegraphics[width=\textwidth]{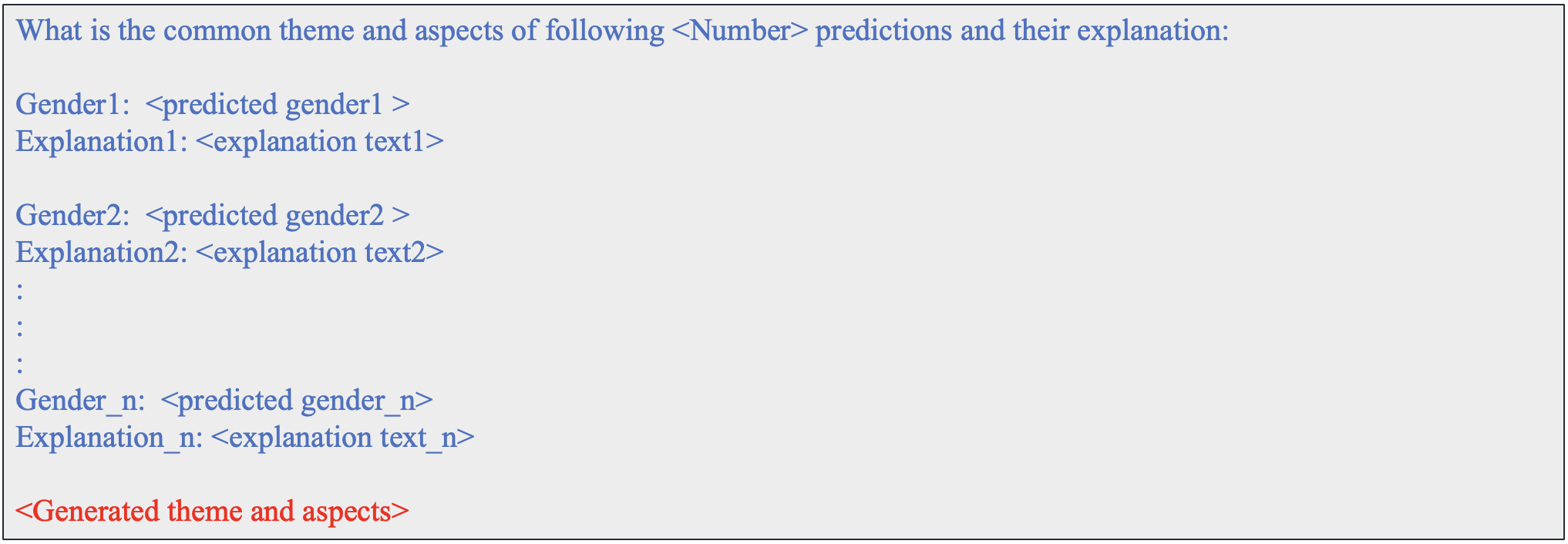}
	\caption{Prompt template for generating theme and aspects from predictions and explanations. Inputs are shown in blue, and outputs are shown in red.}
	\label{thm_gen}
\end{figure*}

Fig. \ref{prompt_tem} illustrates the prompt template.
Fig. \ref{prompt_age} shows the example prompts for age group prediction from the climate campaign dataset. Prompt template for generating theme and aspects from predictions and explanations is shown in Fig. \ref{thm_gen}.
\subsection{Demographic Misclassifications}
\label{app:error}
Table \ref{error} presents an analysis of ad misclassifications based on gender and age group predictions. 
\subsubsection{Misclassification - Senior Instances}
\label{app:senior}
Table \ref{mis_class_senior} shows the details of misclassified senior instances. The misclassification of the Senior age group as Young or Late Working can be attributed to several factors: 
\newline
\textbf{Thematic Content and Topic Association:}
The first three misclassified ads (Table \ref{mis_class_senior}) focus on climate change, environmental activism, and sustainability. These topics are often associated with younger demographics, particularly young adults (18-24), who are perceived to be more engaged in activism and environmental causes. The model appears to have learned an association between these topics and the Young age group, leading to misclassification when seniors engage with similar content.

In Table \ref{mis_class_senior}, the fourth instance shows a misclassification where the model predicts it as a late working (45-64 years) age group. The ad mentions `working families', which is a term commonly associated with individuals in the Late Working age group who are actively engaged in the workforce and supporting families. The content revolves around a political campaign emphasizing the need for change and active participation, themes often associated with the $45-64$ age demographic who are typically more politically active and influential.
\newline
\textbf{Lack of Age-Specific Cues:}
The misclassified ads do not contain explicit references to seniors or age-specific language that would signal the content is intended for the senior demographic. The language is broad and does not mention age-related concerns, such as retirement, health issues prevalent among seniors, or senior-specific programs.
\newline
\textbf{Reliance on Stereotypical Associations:}
The explanations generated by LLMs indicate that the model relies on stereotypes, associating certain topics exclusively with specific age groups. By assuming that environmental activism is primarily of interest to young adults, the model overgeneralizes by overlooking the possibility that seniors are also engaged in these issues.
\newline
\textbf{Feature Representation Limitations:}
The model may lack features that capture subtle cues indicating the ad's target age group when explicit age markers are absent. The model may not effectively utilize contextual information that could hint at the intended audience beyond topic associations. For example, the interests and concerns of the late working and senior age groups can overlap, especially in areas like politics and social change.
\begin{table*}
\centering
\begin{tabular}{p{5cm}|p{2cm}|p{2.5cm}|p{5cm}}
\hline
\textbf{Ad Text} & \textbf{Actual} & \textbf{Prediction (Misclassification)} & \textbf{Explanations by LLMs} \\
\hline
Meet the first carbon negative fragrance. Air Eau de Parfum converts CO2 into something beautiful: sustainable ethanol blended with notes of orange peel, fig leaf, powdery musk and tobacco invigorating scent that redefines what is possible. Available exclusively at aircompany.com & Male & Female & The fragrance notes like fig leaf, powdery musk, and sustainable ethanol typically resonate more with female audiences in fragrance marketing. \\
\hline
Over generations, Illinois farm families have learned how to work with our climate to produce healthy crops and livestock. & Female & Male & The ad emphasizes farming and agriculture, traditionally male-dominated fields, making men the more likely target audience for this message. \\
\hline
Renewable energy is projected to surpass fossil gas as the dominant source of energy production by 2030. Texas, a leader in wind output, is in a great position as the nation transitions. & Young & Early working & This group is likely investing in sustainable energy options, driven by long-term economic and environmental considerations. \\
\hline
\texttt{Take Action Now to Fight Climate Change! Plant a tree in California \url{treestoglobe.org/plantingchallengecalifornia}} & Early working & Young & This age group is often more environmentally active and responds well to social media campaigns promoting climate action initiatives. \\
\hline
The worst impacts of climate change could be irreversible by 2030. The time to switch to a renewable energy plan is now. Not tomorrow. Not next week. & Late working & Early working & They can switch energy plans and are motivated to act before irreversible impacts by 2030. \\
\hline
Tell Rep. Schrader: Now is the time to go big on climate. VOTE YES on the Build Back Better Act. & Senior & Young & Targets young adults passionate about climate action and eager to influence political decisions for their future. \\
\hline
\end{tabular}
\caption{Error Analysis.}
\label{error}
\end{table*}
\begin{table*}
    \centering
    \begin{tabular}{p{6.2cm}|p{2cm}|p{6.2cm}}
        \hline
        \textbf{Ad Text} & \textbf{Prediction} & \textbf{Explanations by LLMs} \\ \hline
        The Hebrew University Center for Climate Science was established in Israel to fight climate change worldwide. & Young &  Targets university-age students interested in climate science to study and combat climate change.\\ \hline
        Tell Rep. Schrader: Now is the time to go big on climate. VOTE YES on the Build Back Better Act. & Young & Targets young adults passionate about climate action and eager to influence political decisions for their future. \\ \hline
        Get clean water and pollution-free electricity to all of America. & Young &  Targets environmentally conscious young adults concerned about sustainability and the future impact of clean water and energy. \\ \hline
        Let’s make one thing clear: Pennsylvania will be the single most competitive Senate race of 2022... It’s one of the reasons I am running for the U.S. Senate seat in Pennsylvania. & Late working &  Targets 45-64-year-olds by focusing on working families, anti-celebrity politics, and referencing the Trump Era. \\ \hline
    \end{tabular}
    \caption{Misclassified Senior Instances.}
    \label{mis_class_senior}
\end{table*}
\subsection{Thematic Analysis of Gender Explanations}
\label{app:thm_gen}
From $1^{st}$ row of Table \ref{gender_thm}, we notice that the overall theme revolves around targeting \textbf{\textit{men}} by aligning ads with their \textbf{perceived interests and roles} in technology, finance, property, traditional male activities, and political or economic discourse. The explanations consistently emphasize the following aspects:
\newline
\textbf{Interest in Technology and Innovation:} Men are often depicted as being more engaged with technology, engineering, and renewable energy solutions. Ads related to technical aspects of engines, energy efficiency, and infrastructure are considered more likely to appeal to men.
\newline
\textbf{Focus on Economic and Financial Issues:} Many explanations suggest that men are more concerned with economic benefits, investment opportunities, and financial savings, making them the likely target for ads that emphasize these aspects.
\newline
\textbf{Property and Land Management:} The theme of land ownership, property value improvement, and land management is frequently mentioned, with the assumption that men are more interested in these areas.
\newline
\textbf{Traditional Male Activities:} Ads that involve traditionally male-oriented activities, such as beer consumption, vehicle-related savings, home maintenance, and physical strength, are seen as more likely to target men.
\newline
\textbf{Engagement in Political and Infrastructure Topics:} Men are often portrayed as more engaged in political discourse, infrastructure initiatives, and discussions around energy and policy, making them the primary audience for ads focused on these themes.
\newline
\textbf{Conservative Views and Skepticism:} Some explanations suggest that men are more likely to resonate with conservative views, skepticism about environmental claims, and anti-establishment sentiments.

From $2^{nd}$ row of Table \ref{gender_thm}, we observe that the overall theme revolves around targeting \textbf{\textit{women}} by aligning ads with their \textbf{roles as caregivers, environmental advocates, and socially conscious individuals} who prioritize the well-being of their families, communities, and the environment. The explanations consistently emphasize the following aspects:
\newline
\textbf{Parental and Caregiving Roles:} Many explanations highlight that women, particularly mothers, are more likely to resonate with messages about protecting children's futures, parental responsibilities, and family well-being. These ads often appeal to maternal instincts and the role of women as primary caregivers.
\newline
\textbf{Environmental Consciousness:} Women are frequently depicted as being more engaged with environmental issues, sustainability, and community health. The explanations suggest that women are more proactive and vocal about climate change, conservation, and eco-friendly initiatives.
\newline
\textbf{Social Welfare and Community Involvement:} The explanations note that women are more likely to be concerned with social issues such as paid leave, affordable childcare, healthcare, and community well-being. Ads that emphasize these themes are seen as more likely to appeal to women.
\newline
\textbf{Empathy and Emotional Appeal:} The explanations often mention that women are more responsive to ads that evoke empathy, emotional concerns, and collective action. This includes ads that focus on protecting the environment for future generations and supporting social safety nets.
\newline
\textbf{Female Empowerment and Leadership:} Some explanations specifically mention themes of women's empowerment, leadership, and support for female scientists or leaders. These themes are likely to resonate more with female audiences who identify with or support gender equality and empowerment.
\newline
\textbf{Health and Safety Concerns:} Women are portrayed as being more attentive to issues related to health, safety, and the well-being of their families and communities. This includes a strong focus on environmental health and sustainability.
\subsection{Thematic Analysis of Age Explanations}
\label{app:thm_age}
From $1^{st}$ row of Table \ref{age_thm}, we observe that the overall theme revolves around the \textbf{activism and the environmental consciousness} of \textit{\textbf{young adults}}, positioning them as a key demographic for campaigns and initiatives focused on climate change and sustainability. The explanations highlight the following aspects:
\newline
\textbf{Passion for Climate Action:} Young adults are described as being particularly passionate about addressing climate change, often leading or participating in environmental activism and campaigns.
\newline
\textbf{Support for Bold Environmental Leadership:} This age group is likely to support bold and urgent actions related to environmental protection and sustainability.
\newline
\textbf{Engagement with Activism:} The explanations emphasize that young adults are more likely to be engaged in climate-related activism and are motivated to take meaningful actions.
\newline
\textbf{Desire for Immediate Change:} There is a recurring mention of the desire for immediate and meaningful change, reflecting the urgency with which young adults approach environmental issues.
\newline
\textbf{Participation in Training and Advocacy:} The group is also characterized as eager to participate in training programs and initiatives that allow them to contribute actively to environmental causes.

From $2^{nd}$ row of Table \ref{age_thm}, we can see that the overall theme revolves around the \textbf{proactive and responsible mindset} of \textbf{\textit{early working-age}} adults, who are not only financially capable but also motivated by a strong sense of social and environmental responsibility. They are seen as key targets for initiatives that combine sustainability with practical, career-oriented, and family-focused benefits. The explanations consistently emphasize the following aspects:
\newline
\textbf{Environmental Consciousness:} This age group is described as being highly engaged with environmental issues, such as climate change, sustainability, and clean energy. They are likely to support initiatives and products that align with eco-friendly values.
\newline
\textbf{Financial Stability and Disposable Income:} Many explanations note that individuals in this group have disposable income, making them financially capable of supporting and investing in sustainable products, services, and causes.
\newline
\textbf{Parental and Future Concerns:} This demographic is often portrayed as parents or future-focused individuals who are concerned about the impact of environmental issues on their children and future generations.
\newline
\textbf{Career Engagement and Professional Roles:} The explanations frequently mention that this age group is active in their careers, often holding decision-making roles that influence corporate and household sustainability practices.
\newline
\textbf{Interest in Innovation and Technology:} Individuals in this age group are also depicted as being interested in innovative industries, clean energy solutions, and sustainability technologies, which align with their professional and personal goals.
\newline
\textbf{Social and Political Engagement:} The group is characterized as being engaged in socio-political issues, particularly those related to corporate accountability, sustainability, and environmental advocacy.

From $3^{rd}$ row of Table \ref{age_thm}, we can notice that the overall theme revolves around the \textbf{responsibilities and concerns} of individuals in the \textbf{\textit{late working (45-64) age}} group, focusing on their roles as homeowners, voters, and economically engaged citizens who are likely to be influenced by environmental, economic, and policy-related messaging. The explanations specifically emphasize the following aspects:
\newline
\textbf{Economic and Environmental Responsibility:} Many of the explanations mention that individuals in this age group are concerned with sustainability, home energy efficiency, and environmental impact. They are likely to invest in public resources and adopt changes that contribute to economic and environmental sustainability.
\newline
\textbf{Homeownership and Financial Stability:} This demographic is characterized as established homeowners who are financially secure. They are seen as key targets for changes related to home energy efficiency, such as adopting solar power, due to their financial means and homeownership status.
\newline
\textbf{Voter and Policy Engagement:} The explanations suggest that this age group is politically active, particularly concerned with public safety, and likely to support policy changes by voting on local measures.
\newline
\textbf{Economic Concerns:} There is an emphasis on economic factors such as unemployment, inflation, and gas prices, with concerns about current economic policies affecting their businesses and financial stability.

From $4^{th}$ row of Table \ref{age_thm}, we notice that the overall theme centers on \textbf{health and safety concerns} that are particularly important to \textbf{\textit{senior citizens}}, with a focus on programs that cater to their specific needs and the heightened risks they face in certain situations. The key aspects highlighted in the explanations are:
\newline
\textbf{Health and Wellness Programs:} The first explanation mentions programs like SilverSneakers and Silver\&Fit, which are specifically designed for senior citizens to support their physical health and well-being.
\newline
\textbf{Vulnerability and Safety:} The second explanation focuses on the increased vulnerability of seniors to COVID-19 and emphasizes the risks they face, particularly in the context of political decisions or public health issues.

\subsection{Unsupervised Baselines}
\label{unsup_base}
\begin{figure*}
    \centering
	\includegraphics[width=\textwidth]{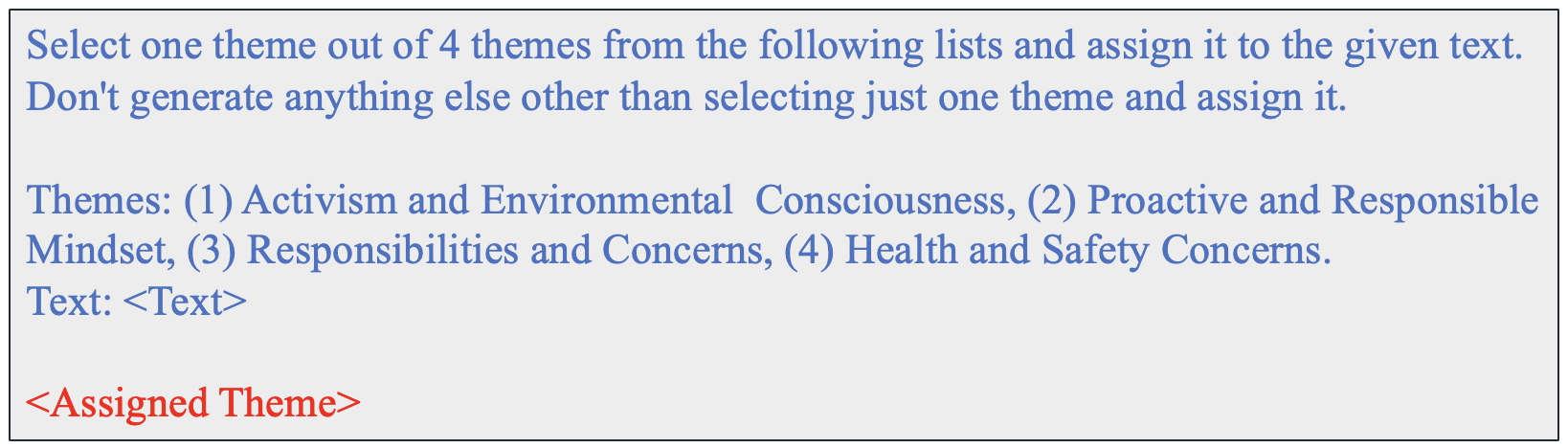}
	\caption{Prompt template for assigning theme using LLM. Inputs are shown in blue, and outputs are shown in red.}
	\label{assign_thm_llm}
\end{figure*}

Generated topics from the prediction explanation
of gender and age using LDA and BERTopic is shown in Table \ref{tab:topic_modeling_results}. Table \ref{tab:human} shows the human evaluation results on GPT-4o assignment, LDA, and BERTopic. Prompt template for LLM assignment is shown in Fig. \ref{assign_thm_llm}.

\begin{table}[h!]
\centering
\begin{tabular}{l|l|c}
\toprule
\textbf{Model} & \textbf{Demo.} & \textbf{Acc. (\%)} \\
\midrule
\multirow{2}{*}{LDA} & gender & 61.32 \\
                     & age    & 33.06     \\
\hline
\multirow{2}{*}{BERTopic} & gender & 56.60 \\
                          & age    & 45.45     \\
\hline
\multirow{2}{*}{GPT-4o} & gender & $\textbf{88.67}$ \\
& age   & $\textbf{68.56}$ \\
\bottomrule
\end{tabular}
\caption{Assignment Comparison w.r.t. Human Judgments. Demo.: Demographics, Acc.: Accuracy.}
\label{tab:human}
\end{table}
\begin{table*}
\centering
\begin{tabular}{l|l|p{11cm}}
\hline
\textbf{Dimension} & \textbf{Method} & \textbf{Topics / Top Words} \\
\hline
\multirow{10}{*}{Gender} 
& \multirow{4}{*}{BERTopic} & \textbf{Topic 1}: Practical and Economic Appeals to Men (Technology, Property, Costs) \\
&          & \textbf{Topic 2}: Emotional and Social Appeals to Women (Spirituality, Motherhood, Environment) \\
\cline{2-3}
& \multirow{6}{*}{LDA}      & \textbf{Topic 1}: like, resonate, men, ads, males, focus, property, care, leave, paid \\
&          & \textbf{Topic 2}: men, ad, male, energy, traditionally, engagement, likely, emphasizes, climate, higher \\
&          & \textbf{Topic 3}: ad, female, environmental, emphasizes, resonate, typically, males, audiences, target, ads \\
&          & \textbf{Topic 4}: women, ad, environmental, making, issues, future, likely, mothers, children, concerned \\
\hline
\multirow{10}{*}{Age} 
& \multirow{4}{*}{BERTopic} & \textbf{Topic 1}: Career-Oriented Adults Focused on Sustainability and Economic Impact \\
&          & \textbf{Topic 2}: Young, Socially Active, and Environmentally Conscious Professionals \\
\cline{2-3}
& \multirow{6}{*}{LDA}       & \textbf{Topic 1}: climate, action, ad, likely, change, young, passionate, targets, eager, adults \\
&          & \textbf{Topic 2}: likely, engaged, environmental, group, energy, climate, issues, adults, support, typically \\
&          & \textbf{Topic 3}: group, likely, age, sustainability, energy, aligning, engaged, clean, career, corporate \\
&          & \textbf{Topic 4}: group, conscious, environmentally, sustainable, making, disposable, income, likely, like, products \\
\hline
\end{tabular}
\caption{Comparison of BERTopic and LDA-generated topics for demographic dimensions of Gender and Age. BERTopic provides interpretable topic labels, while LDA results include top-10 words per topic.}
\label{tab:topic_modeling_results}
\end{table*}
\subsection{Sensitivity Analysis on Fairness}
\label{senstv_ana}
\begin{figure*}[htbp]
\centering
\begin{subfigure}{\columnwidth}
  \centering
  \includegraphics[width=\textwidth]{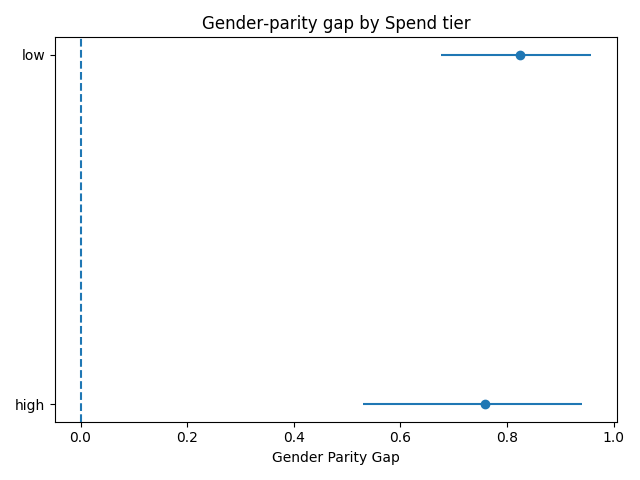}
  \caption{Fairness by spend tier (Gender).}\label{fig:gps}
\end{subfigure}%
\begin{subfigure}{\columnwidth}
  \centering
  \includegraphics[width=\textwidth]{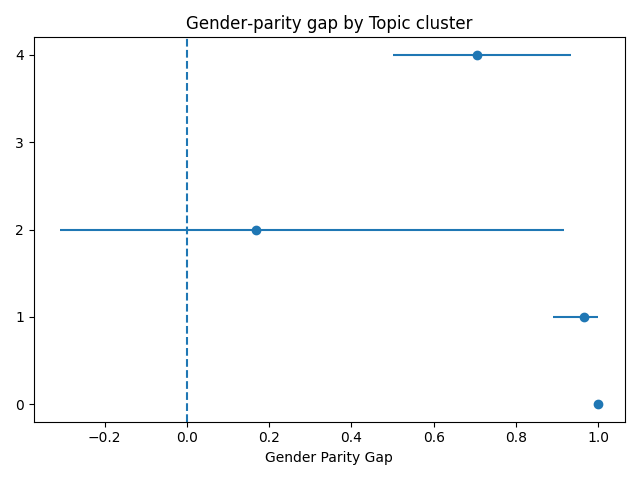}
  \caption{Fairness by topic cluster (Gender).}\label{fig:gpt}
\end{subfigure}
\begin{subfigure}{\columnwidth}
  \centering
  \includegraphics[width=\textwidth]{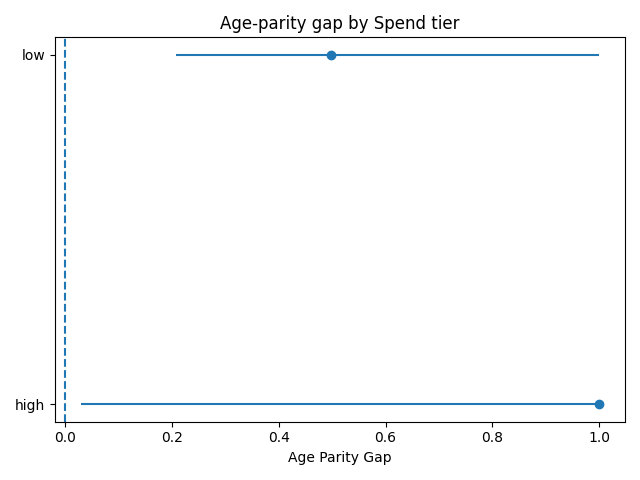}
  \caption{Fairness by spend tier (Age)}\label{fig:aps}
\end{subfigure}%
\begin{subfigure}{\columnwidth}
  \centering
  \includegraphics[width=\textwidth]{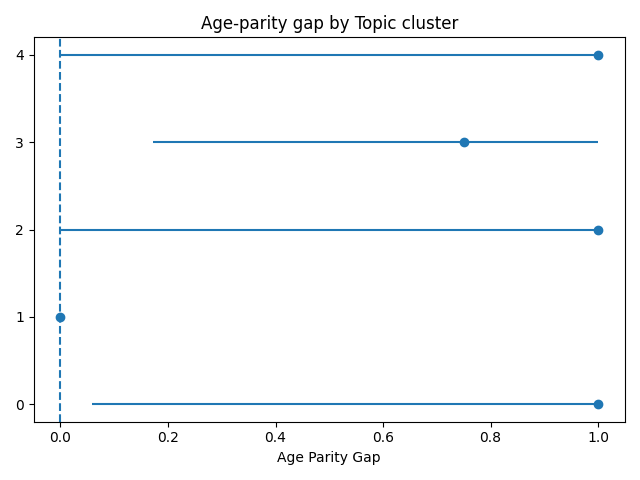}
  \caption{Fairness by topic cluster (Age).}\label{fig:apt}
\end{subfigure}
\caption{Sensitivity Analysis: Rerunning the fairness test after stratifying by ad-spend and by topic cluster.}
\label{fig:sense}
\end{figure*}

We explicitly frame our work as a \textbf{post-hoc audit} and the manuscript avoids causal verbs such as “impact” or “cause” when reporting accuracy or fairness metrics. Fig. \ref{fig:sense} shows the sensitivity analysis by rerunning the fairness test after stratifying by ad-spend and by topic cluster. 

\subsection{Learning-Curve Robustness Analysis}
\label{lc_robust}
\begin{figure*}[htbp]
\centering
\begin{subfigure}{\columnwidth}
  \centering
  \includegraphics[width=\textwidth]{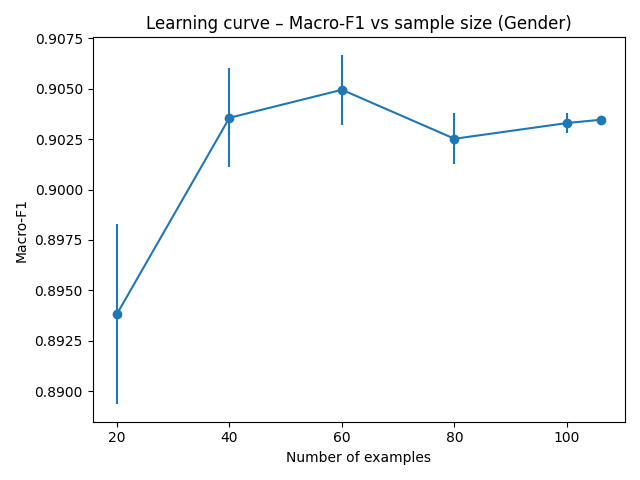}
  \caption{F1 vs sample size (Gender).}\label{fig:lc_f1_gen}
\end{subfigure}%
\begin{subfigure}{\columnwidth}
  \centering
  \includegraphics[width=\textwidth]{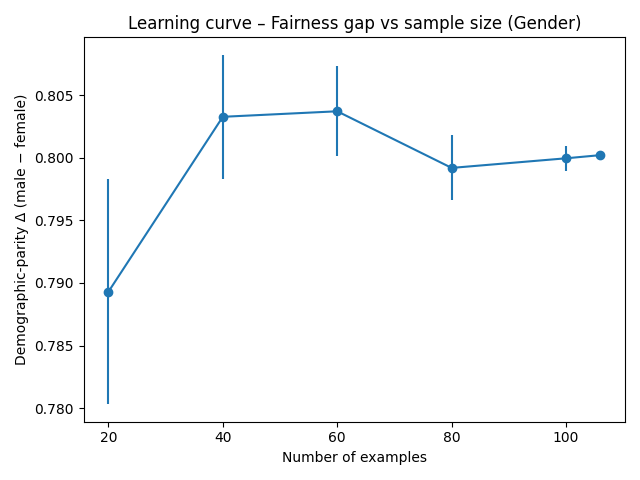}
  \caption{Parity gap vs sample size (Gender).}\label{fig:lc_fair_gen}
\end{subfigure}
\begin{subfigure}{\columnwidth}
  \centering
  \includegraphics[width=\textwidth]{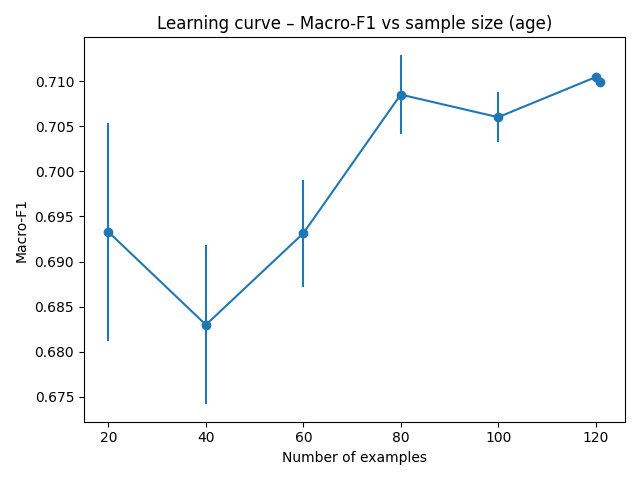}
  \caption{F1 vs sample size (Age).}\label{fig:lc_f1_age}
\end{subfigure}%
\begin{subfigure}{\columnwidth}
  \centering
  \includegraphics[width=\textwidth]{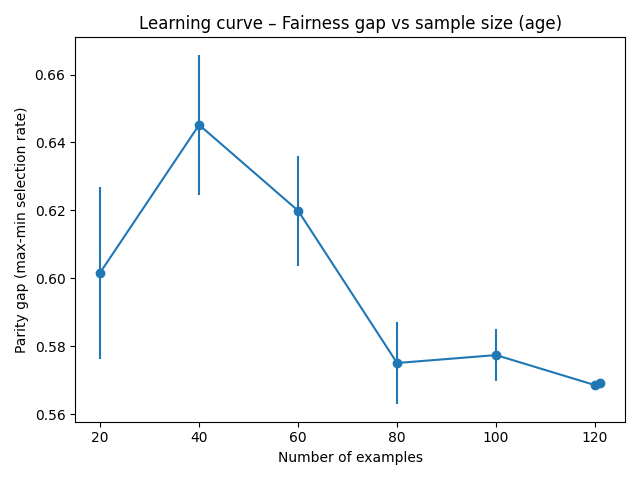}
  \caption{Parity gap vs sample size (Age).}\label{fig:lc_fair_age}
\end{subfigure}
\caption{Learning-curve robustness analysis.}
\label{fig:lcr}
\end{figure*}

Fig. \ref{fig:lcr} shows the learning curve having macro average F1 score vs. sample size and fairness gaps vs. sample size.




\end{document}